\documentclass{article}

\usepackage[final]{neurips_2024}

\usepackage[utf8]{inputenc} 
\usepackage[T1]{fontenc}    
\usepackage{hyperref}       
\usepackage{url}            
\usepackage{booktabs}       
\usepackage{amsfonts}       
\usepackage{nicefrac}       
\usepackage{microtype}      
\usepackage{xcolor}         
\usepackage[colorinlistoftodos]{todonotes}
\usepackage{xspace}
\usepackage{siunitx}
\newcommand{\ours}{SALT}
\usepackage[yyyymmdd]{datetime}

\usepackage{amsmath}
\usepackage{amssymb}
\usepackage{arydshln}
\usepackage{makecell}
\usepackage{graphicx}
\usepackage{caption}
\usepackage{subcaption}
\usepackage{multirow}
\usepackage{enumitem}
\usepackage{svg}
\usepackage[symbol]{footmisc}
\svgpath{{./figs/}} 

\usepackage{calc} 
\usepackage{booktabs} 
\usepackage{listings}
\newsavebox{\mybox}

\title{SALT: Sales Autocompletion Linked Business Tables Dataset}

\author{
Tassilo Klein \thanks{Corresponding author: tassilo.klein@sap.com}
\And
Clemens Biehl
\And
Margarida Costa
\And
Andre Sres
\And
Jonas Kolk
\And
Johannes Hoffart \AND SAP SE
}

\begin{document}

\maketitle

\begin{abstract}
Foundation models, particularly those that incorporate Transformer architectures, have demonstrated exceptional performance in domains such as natural language processing and image processing. Adapting these models to structured data, like tables, however, introduces significant challenges. These difficulties are even more pronounced when addressing multi-table data linked via foreign key, which is prevalent in the enterprise realm and crucial for empowering business use cases. Despite its substantial impact, research focusing on such linked business tables within enterprise settings remains a significantly important yet underexplored domain.
To address this, we introduce a curated dataset sourced from an Enterprise Resource Planning (ERP) system, featuring extensive linked tables. This dataset is specifically designed to support research endeavors in table representation learning. By providing access to authentic enterprise data, our goal is to potentially enhance the effectiveness and applicability of models for real-world business contexts.\footnote[7]{Data and code available at \url{https://github.com/sap-samples/SALT}}
\end{abstract}

\section{Introduction}
Deep learning has made substantial strides in areas like text understanding, language translation, image classification, and object detection. These advancements are largely driven by foundational models trained on diverse datasets and self-supervised training techniques, especially those that incorporate Transformer architectures. However, using these models on structured, tabular data, essential for enterprise business operations, poses unique challenges. These challenges become more pronounced with multi-table configurations consisting of large tables interconnected by foreign keys and comprising extensive business datasets, a setup to which we refer to as \emph{linked business tables}. Such setups are common in real-world business scenarios.
The challenges in applying foundational models to linked business data are primarily twofold: algorithmic and data-related. Algorithmically, a significant challenge is adapting models that were originally designed for unstructured internet data to handle structured data effectively - see~\citep{grinsztajn2022why} for a comprehensive discussion. This process requires a sophisticated integration of structural knowledge and the unique characteristics of linked business data, which is inherently more complex and interconnected than straightforward internet-scraped table data. \\
One major limitation in the current landscape is the absence of realistic, enterprise-linked multi-table datasets at scale. Existing table datasets often originate from HTML pages and do not accurately represent the complexity and dynamics of expansive database tables used in active enterprise systems~\citep{BodensohnBrackmannVogel2024}. Moreover, obtaining large, clean, and high-quality datasets for structured tabular applications presents difficulties~\citep{hulsebos2023gittables,pmlr-v235-van-breugel24a}, particularly in enterprise settings where data privacy, confidentiality, and commercial interests restrict data access. This lack of suitable public datasets leads to significant domain adaptation challenges and shifts in data distribution, which pose difficulties for many existing models~\citep{pmlr-v235-fey24a}.
To tackle these issues, we have curated the Sales Autocompletion Linked Business Tables (SALT) dataset, sourced from an Enterprise Resource Planning (ERP) system. ERP systems are comprehensive, multifunctional platforms essential for managing all core business operations including finance, human resources, production, and supply chains. As the backbone of organizational data management, ERP systems provide an excellent foundation for developing and evaluating data models that accurately reflect complex, real-world enterprise environments. The SALT dataset, which includes interconnected relational tables with a focus on sales, encompasses several million entries across various enterprise sales operations (cf. synthetic sales dataset SalesDB~\cite{motl2024ctupraguerelationallearning}).
By sharing the SALT dataset with the research community, we aim to stimulate advancements in table representation learning and refine algorithm development to enhance applicability and performance in real-world settings. This initiative is crucial for evolving deep learning models that not only understand but also effectively function within the complexities of large-scale enterprise data landscapes, thereby promoting the development of enterprise-specific machine learning applications.

\noindent{\textbf{Related Work: }} The majority of existing table datasets originate from scraping the Web, notably extracted from HTML pages or CSV files from GitHub, which inadequately capture the complexity and dynamics typical of large database tables that are employed in operational enterprise systems.
WebTables~\citep{10.14778/1453856.1453916} corpus includes a massive collection of 233 million tables, sourced from HTML pages via the Common Crawl project. While WebTables offers an extensive quantity of tables, its diversity is constrained because it solely comprises HTML tables from web pages. TURL \cite{10.14778/3430915.3430921} provides a cleaner corpus of 580 thousand tables extracted from Wikipedia.
In contrast, GitTables~\citep{hulsebos2023gittables} contains over 10 million tables extracted from "comma-separated value" files (CSVs) found on GitHub. Tables from GitTables generally exhibit structural differences compared to those from WebTables, making GitTables an essential corpus despite its focus on a single file type.
TabLib~\citep{eggert2023tablib} comprises 627 million tables across various file formats and totaling 69 TiB, sourced from GitHub and Common Crawl. Notably, it comprises exceptionally large tables of several million rows and columns.
LakeBench~\citep{LakeBench2024} is a collection of benchmarks to resemble enterprise data lakes, containing tables from a variety of
sources such as open government data for the purpose of unionability, joinability, and subset tasks.

\begin{figure*}
    \centering
\includegraphics[width=0.80\textwidth]{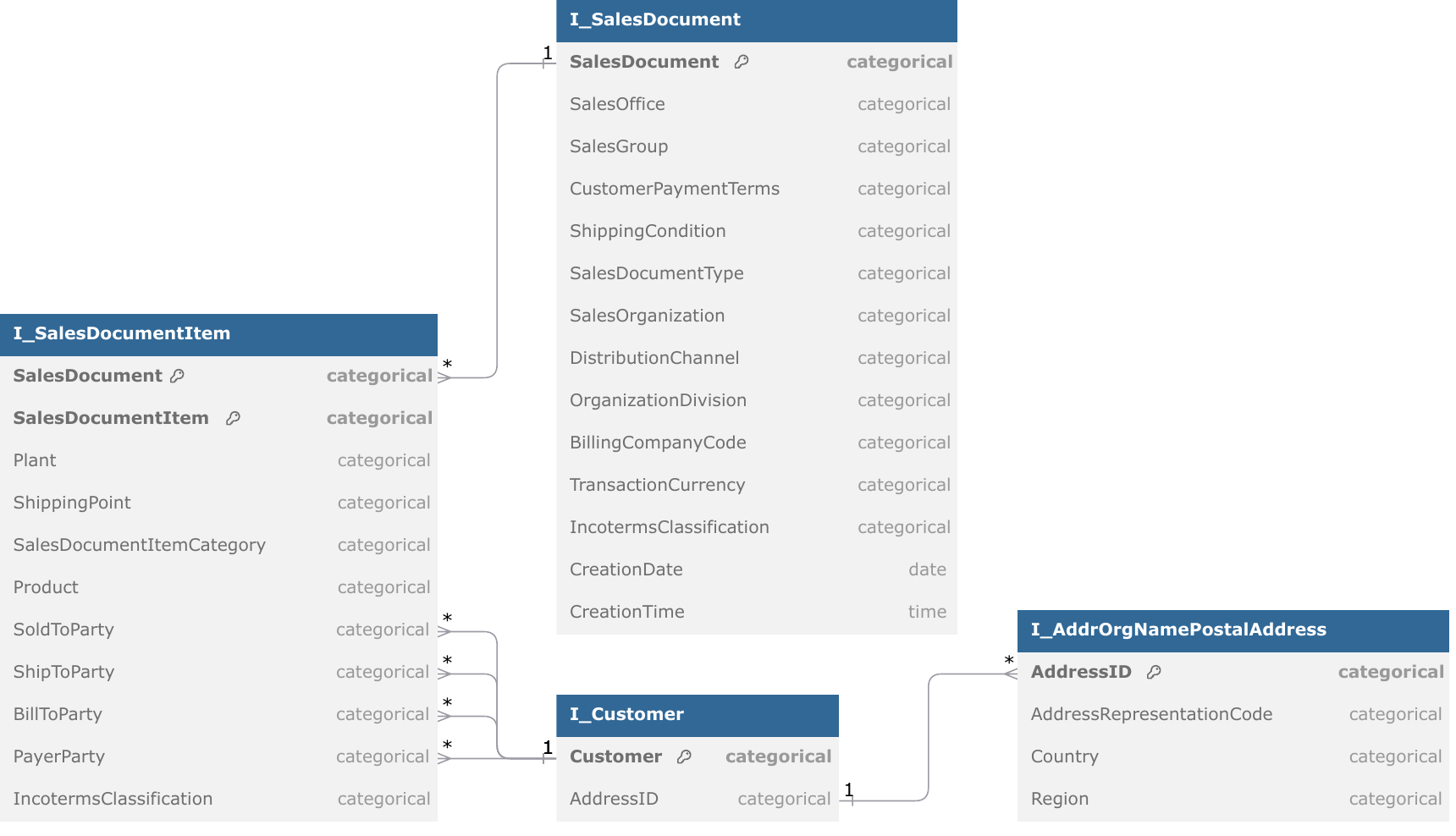}
    \caption{\textbf{Dataset Schemas}. Schemas for the four tables constituting the \ours~dataset. Primary keys are highlighted in \textbf{bold} letters and with a key symbol. Foreign keys interconnecting tables. }
    \label{fig:schema}
     \vspace{-3mm}
\end{figure*}

\section{SALT Dataset}
\label{sec:salt-dataset}
\noindent{\textbf{Background: }}
The SALT dataset is specifically curated to mirror customer interactions within an Enterprise Resource Planning (ERP) system and is designed to train models that assist users by predicting fields typically missing in sales orders - see Fig.~\ref{fig:s4hanasalesorderscreenshot} in the Appendix for screenshots of the user interface of a sales order application. This dataset is crucial to the sales and distribution process, especially for creating the "Sales Order Document." Each of these documents records a single transaction that includes various items, marking a distinct phase in the sales cycle. \\
Structured around four principal tables—sales documents, sales document items, customers, and addresses—the dataset consolidates data from a single enterprise that underwent anonymization (for details see Appendix Sec.~\ref{sec:anonymization}). The sales documents table logs vital details such as sales office, sales group, payment conditions, and shipping arrangements, limiting its entries to those specifically categorized as sales orders. The sales document items table captures detailed information for each line item in these documents, including the product sold, the shipping point, and the parties involved in the transaction. Concurrently, the customer table holds comprehensive master data about customers, further elaborated in the addresses table with specifics like country and region. The input variables in the dataset include a mix of fields typically populated by users during the creation of a sales order, augmented by master data fields like material number and customer details. The target variables are not always maintained; they are optional and may not be filled out for certain transactions depending on particular scenarios or requirements. This intricate structure of SALT not only enhances model training for missing field predictions but also effectively replicates complex ERP interactions.
\begin{table*}\centering
\caption{\textbf{Breakdown of atomic data composition of datasets:} Datasets \ours, \emph{GitTables}~\citep{hulsebos2023gittables}, \emph{WebTables}~\citep{2016_lehmberg_wdc_corpus} and \emph{TabLib}~\citep{eggert2023tablib}. Results except for ours are taken from the respective papers.}
\label{tab:atomic-composition}
\begin{tabular}{lccccc}
   \toprule
\textbf{Atomatic data type}               & \textbf{\ours}        & \textbf{GitTables}          & \textbf{WebTables}      & \textbf{TabLib}    \\
    \midrule
Numeric & 58.1\% & 57.9\% &  51.4\% & 33.6\%\\
String & 38.7\% & 41.6\% &  47.4\% &61.8\%\\
Other & 3.2\% & 0.5\% &  1.2\% &4.6\%\\
\bottomrule
\end{tabular}
\end{table*}

\noindent{\textbf{Task: }}
In the dataset, 21 fields are categorized as potential input variables, serving as features for predictive modeling applications, while 8 fields are designated as target variables, intended for prediction based on the input data analysis. The predictive model, which will be trained using this dataset, is specifically tasked with performing multiclass classification on seven critical variables. These variables are essential for ensuring the seamless execution of sales orders:

\setlist{nolistsep}
\begin{itemize}[noitemsep]
    \item\texttt{I\_SalesDocument.SalesOffice} -  Sales activities for specific products and regions
     \item\texttt{I\_SalesDocument.SalesGroup} - Subdivisions of a distribution chain
     \item\texttt{I\_SalesDocument.CustomerPaymentTerms} - Payment conditions, i.e., deadlines and early payment discounts
     \item\texttt{I\_SalesDocument.ShippingCondition} - Logistics terms
\item\texttt{I\_SalesDocumentItem.ShippingPoint} - Dispatch location
     \item\texttt{I\_SalesDocumentItem.Plant} - Production/ storage facility, critical for inventory control
     \item\texttt{I\_SalesDocument.IncotermsClassification} and \\ \texttt{I\_SalesDocumentItem.IncotermsClassification} - International commercial terms, outline transaction responsibilities like shipping and insurance\footnote{This field can be defined independently on item and header level, which is why both are included.}
\end{itemize}

\noindent{\textbf{Structure: }} The dataset is structured into four primary tables encompassing a total of 500,908 sales orders (\texttt{I\_SalesDocument}), which include 2,319,944 sales order items (\texttt{I\_SalesDocumentItem}) associated with 139,611 unique business partners (\texttt{I\_Customer}) and 1,788,887 (\texttt{I\_AddrOrgNamePostalAddress}) addresses - see Fig.~\ref{fig:schema} for the table schemas. The table fields are filtered to include only the data relevant to the specific use case described above.
After filtering, the tables are merged to form a single flat dataset containing 2,319,944 rows, such that each row in the dataset represents a single sales order item (for details see Appendix Sec. \ref{sec:data_extraction}). The entries cover transactions conducted between January 1, 2018, and December 31, 2020. To assess the dataset's predictive modeling utility, data was divided into temporal splits, with validation segments starting from February 1, 2020, and test segments from July 1, 2020. For an analysis of the distribution values, see Tab.~\ref{tab:atomic-composition}.


\noindent{\textbf{Data Insights:}}
The dataset employed in this study is derived from authentic industry data captured by an Enterprise Resource Planning (ERP) system, documenting sales orders.  This dataset has undergone minimal pre-processing primarily aimed at addressing privacy concerns. Several challenges arise from the nature and quality of the dataset, which need careful consideration:
\setlist{nolistsep}
\begin{itemize}[noitemsep]
\item \textbf{Diversity:} There is a substantial diversity in certain data fields due to the wide range of unique values they contain. For instance, the field \texttt{I\_SalesDocumentItem.ShipToParty} includes 18,097 distinct customer IDs, while \texttt{I\_SalesDocumentItem.Product} comprises 187,562 unique product identifiers.
\item \textbf{Class imbalance:} The dataset demonstrates a pronounced class imbalance. The distribution of sales offices across sales orders is highly skewed; the most frequently occurring sales office is associated with over 60\% of the orders, and the two most common sales offices collectively account for 99\% of the data. Despite this, there are 41 distinct sales offices represented in more than one order, suggesting a long-tail distribution.
\item \textbf{Noise:} A considerable amount of input noise is evident within the dataset. Since data entry is frequently manual, discrepancies may arise as different employees might handle identical business scenarios differently or make inadvertent errors. Moreover, certain fields may be occasionally left blank, potentially leading to gaps in the data.
\item \textbf{Data drift:} Technically, the dataset is prone to data drift, a phenomenon where the categorizations, such as sales groups within the ERP system, evolve over time. This drift may particularly impact analyses involving temporal splits of the data, as category definitions may shift across the time periods. Notably, the target categories are not subject to such drift.
\end{itemize}

\begin{table*}[ht!] \centering
\caption{\textbf{Classification performance of baseline models:} Evaluation of baseline models on the eight different tasks on \ours. \textbf{Top:} Simple baselines \textbf{Middle:} Gradient-boosted decision tree models \textbf{Bottom:} Deep learning methods. \textbf{Performance metric:} Mean Reciprocal Rank.}
\label{tab:classification-performance}
 \resizebox{12.0cm}{!}{%
\begin{tabular}{lccccc}
   \toprule
        \multicolumn{6}{c}{\it{Performance Baseline - MRR~$(\uparrow)$}}\\
         \midrule
\textbf{Method \textbackslash~Target Variable} &\multicolumn{1}{p{3cm}}{\centering  \textbf{Plant} } & \multicolumn{1}{p{3cm}}{\centering  \textbf{Shipping Point} } & \multicolumn{1}{p{3cm}}{\centering  \textbf{Item Incoterm Cls.} } &\multicolumn{1}{p{3.5cm}}{\centering  \textbf{Header Incoterm  Cls.}}\\
    \midrule
Random Classifier&0.59&0.11&0.62&0.62\\
Majority Class Baseline &0.59&0.54&0.62&0.62\\
\hdashline
XGBoost~\citep{Chen:2016:XST:2939672.2939785} &\textbf{0.99}&0.95&0.70&0.70\\
LightGBM~\citep{10.5555/3294996.3295074}&0.61&0.28&0.73&0.73&\\
CatBoost~\citep{Catboost2021}&\textbf{0.99}&0.80&\textbf{0.80}&\textbf{0.81}&\\
\hdashline
CARTE~\citep{CARTE}&\textbf{0.99}&0.97&0.75&0.77\\
AutoGluon~\citep{agtabular} &\textbf{0.99}&0.98&0.78&0.78 &\\
GraphSAGE~\citep{GraphSAGE}&\textbf{0.99}&0.97&0.64&0.59\\
\bottomrule
\\
 \textbf{(continued)} & \multicolumn{1}{p{3cm}}{\centering  \textbf{Sales Office} }   & \multicolumn{1}{p{3cm}}{\centering  \textbf{Sales Group} } & \multicolumn{1}{p{3cm}}{\centering  \textbf{Pay. Terms} } & \multicolumn{1}{p{3cm}}{\centering  \textbf{Ship. Condition} } & \multicolumn{1}{p{1cm}}{\centering  \textbf{Avg.} } \\
    \midrule
Random Classifier&\textbf{0.99}&0.02&0.02&0.12&0.39\\
Majority Class Baseline&\textbf{0.99}&0.05&0.23&0.41&0.51\\
\hdashline
XGBoost~\citep{Chen:2016:XST:2939672.2939785}&\textbf{0.99}&0.51&0.57&0.68&0.76\\
LightGBM~\citep{10.5555/3294996.3295074}&\textbf{0.99}&0.02&0.10&0.51&0.50\\
CatBoost~\citep{Catboost2021}&\textbf{0.99}&0.16&0.44&0.71&0.71\\
\hdashline
CARTE~\citep{CARTE}&\textbf{0.99}&0.46&0.62&0.74&\textbf{0.79}\\
AutoGluon~\citep{agtabular}&\textbf{0.99}&0.34&0.52&0.74&0.77\\
GraphSAGE~\citep{GraphSAGE}&\textbf{0.99}&0.20&0.39&0.59&0.67\\
\bottomrule
\end{tabular}}
\vspace{-2mm}
\end{table*}

\section{Experiments \& Results}
We evaluate the SALT dataset using several baselines for tabular data on the joined table, except for GraphSAGE~\citep{GraphSAGE}, which operates natively in a multi-table setup. The only preprocessing applied is filling in missing values with either a constant value (for the categorical features) or the mean value (for numerical features). The fields related to creation date and time were only used to split the data and then discarded. The validation set was used for early stopping and no hyperparameter tuning was performed.
See Tab.~\ref{tab:classification-performance} for the detailed breakdown of performance evaluation tasks of each task. As can be seen, the Carte ~\citep{CARTE}  shows the best performance on SALT with a margin of $(+0.02~p.)$. The next best approach is AutoGluon ~\citep{agtabular}, followed closely by XGBoost~\citep{Chen:2016:XST:2939672.2939785}.
The analysis reveals several noteworthy insights:
\textbf{i)} Certain target variables demonstrate substantial predictability, achieving prediction scores near 0.99, indicating a high degree of accuracy. \textbf{ii)} The dataset exhibits significant class imbalance, which is particularly evident from the performance of the majority class baseline. This imbalance is most pronounced when predicting variables such as the Sales Office. \textbf{iii)} The predictive performance of the model is adversely affected when tasked with predicting fields like the Sales Group, which suffers from high cardinality issues.

\section{Conclusion}
We introduce a novel dataset focused on linked business data, demonstrating the characteristics of data within actual enterprise systems. We further assessed the performance of current tabular models against tree-based and cutting-edge models. The empirical data reveal that most tabular models effectively manage the prediction tasks in SALT. To augment the dataset's complexity and utility in future work, we plan to include additional tables from a broader range of scenarios, data from multiple companies, and enhance the semantic richness of the dataset to present greater challenges.

\bibliography{refs}
\bibliographystyle{acl_natbib}

\appendix

\section{Appendix}

\subsection{Privacy \& Anonymization}
\label{sec:anonymization}
In adherence to established best practices, our study systematically purged all personally identifiable, company-identifying information, and confidential information from the dataset through a composite of automated and manual processes, thereby mitigating any privacy-related issues. For further assurance of privacy preservation, the sanitized data consequentially underwent a meticulous auditing process. The resulting sanitized data referred to as the \ours{}, comprised exclusively encrypted categorical variables. It is noteworthy to emphasize that our privacy sanitization protocol was designed to preclude any likelihood of data distribution distortion or introduction of bias.

\subsection{Table Detailed Information}
This section describes the schemas of the four tables that constitute the \ours~ dataset. Figure~\ref{fig:schema} shows the schema of the tables. See Tab.~\ref{tab:sales-document-item} for a detailed overview of the \emph{SalesDocumentItem}, Tab.~\ref{tab:sales-document} for \emph{SalesOrder} details, Tab.~\ref{tab:customer} for customer data detail and Tab.~\ref{tab:address} for the customer address details. The column \textit{Is Target Field} indicates the target fields that should be predicted based on the other fields in the tables since they are populated in a later stage of the sales order creation.
\subsection{Additional Data Statistics}
This section provides additional statistics on data composition. Table~\ref{tab:target-field-stats} provides statistics on the target fields.
Table~\ref{tab:target-field-stats} provides detailed statistics of the table fields.

\begin{table}[ht!]
\centering
\caption{\textbf{SalesDocument in Detail}}
\vspace{1.5mm}
\resizebox{14cm}{!}{%
\begin{tabular}{|c|c|c|c|}
\hline
\thead{Field Name} & \thead{Data Type} & \thead{Description} & \thead{Is Target Field} \\ \hline
SalesDocument & Categorical (integer) & ID of the sales document &  \\ \hline
SalesDocumentType  & Categorical (string)  & Type of sales document & \\ \hline
SalesOrganization & Categorical (string)& ID of the sales organization &  \\ \hline
DistributionChannel & Categorical (string)& ID of the distribution channel &  \\ \hline
OrganizationDivision & Categorical (string)& ID of the organization division &  \\ \hline
BillingCompanyCode & Categorical (string) & Company code to be billed & \\ \hline
TransactionCurrency & Categorical (string) & Currency code (EUR, USD, \dots) & \\ \hline
CreationDate & Date (string) & Date of sales document creation & \\ \hline
CreationTime & Time (string) & Time of day of sales document creation & \\ \hline
SalesOffice & Categorical (string)& ID of the sales office & \checkmark \\ \hline
SalesGroup & Categorical (string)& ID of the sales group & \checkmark \\ \hline
CustomerPaymentTerms & Categorical (string) & ID of the payment terms & \checkmark\\ \hline
ShippingCondition & Categorical (string) & ID of the shipping condition & \checkmark\\ \hline
IncotermsClassification & Categorical (string) & ID of the incoterms & \checkmark \\ \hline
\end{tabular}}
\label{tab:sales-document}
\end{table}

\begin{table}[ht!]
\centering
\caption{\textbf{AddrOrgNamePostalAddress in Detail}}
\resizebox{14cm}{!}{%
\begin{tabular}{|c|c|c|c|}
\hline
\thead{Field Name} & \thead{Data Type} & \thead{Description} & \thead{Is Target Field} \\ \hline
AddressID & Categorical (integer)& ID of the address &  \\ \hline
AddressRepresentationCode & Categorical (integer) & System internal address code &  \\ \hline
Country & Categorical (string) & Country name &  \\ \hline
Region & Categorical (string) & Region name &  \\ \hline
\end{tabular}}
\label{tab:address}
\end{table}

\begin{table}[ht!]
\centering
\caption{\textbf{SalesDocumentItem in Detail}}
\resizebox{14cm}{!}{%
\begin{tabular}{|c|c|c|c|}
\hline
\thead{Field Name} & \thead{Data Type} & \thead{Description} & \thead{Is Target Field} \\ \hline
SalesDocument & Categorical (integer) & ID of the sales document &  \\ \hline
SalesDocumentItem & Categorical (integer) & ID of the sales document item &  \\ \hline
SalesDocumentItemCategory & Categorical (string)& ID of the item category &  \\ \hline
Product & Categorical (integer)& ID of the product sold &  \\ \hline
SoldToParty & Categorical (integer)& ID of the customer sold to &  \\ \hline
ShipToParty & Categorical (integer)& ID of the customer shipped to &  \\ \hline
BillToParty & Categorical (integer)& ID of the customer billed to &  \\ \hline
PayerParty & Categorical (integer)& ID of the payer &  \\ \hline
Plant & Categorical (string) & ID of the plant & \checkmark \\ \hline
ShippingPoint & Categorical (string)& ID of the shipping point & \checkmark \\ \hline
IncotermsClassification & Categorical (string) & ID of the incoterms & \checkmark \\ \hline
\end{tabular}}
\label{tab:sales-document-item}
\end{table}

\begin{table}[ht!]
\centering
\caption{\textbf{Customer in Detail}}
\begin{tabular}{|c|c|c|c|}
\hline
\thead{Field Name} & \thead{Data Type} & \thead{Description} & \thead{Is Target Field} \\ \hline
Customer & Categorical (integer)& ID of the customer &  \\ \hline
AddressID & Categorical (integer)& ID of customer's address & \\ \hline
\end{tabular}
\label{tab:customer}
\end{table}

\begin{table*}[ht!]
\centering
\caption{\textbf{Target field statistics}\\
}
\vspace{1.5mm}
\resizebox{14cm}{!}{%
\begin{tabular}{lccc}
   \toprule
\textbf{Target field}               & \textbf{Unique values (\#)}        & \textbf{Missing values (\%)}          & \textbf{Normalized entropy}          \\
    \midrule
I\_SalesDocument.CustomerPaymentTerms & 158 & 1.09 &  0.53\\
I\_SalesDocument.IncotermsClassification & 14 & 0.66 &  0.51\\
I\_SalesDocument.SalesGroup & 589 & 0.82 &  0.84\\
I\_SalesDocument.SalesOffice & 45 & 0.01 &  0.18\\
I\_SalesDocument.ShippingCondition & 56 & 0.01 &  0.55\\
I\_SalesDocumentItem.IncotermsClassification & 14 & 0.24 &  0.5\\
I\_SalesDocumentItem.Plant & 39 & 0.02 &  0.46\\
I\_SalesDocumentItem.ShippingPoint & 97 & 0.03 &  0.48\\
\bottomrule
\end{tabular}}
\label{tab:target-field-stats}
\end{table*}

\begin{table*}[ht!]\centering
\caption{\textbf{Field statistics:} Number of unique values, percentage of missing values, normalized entropy.}
\vspace{1mm}
\resizebox{14cm}{!}{%
\begin{tabular}{llccc}
   \toprule
\textbf{Table} & \textbf{Field}            & \textbf{Unique values (\#)}        & \textbf{Missing values (\%)}          & \textbf{Normalized entropy}          \\
    \midrule
I\_SalesDocument & BillingCompanyCode & 31 & 0.0 & 0.59 \\
I\_SalesDocument & CreationDate & 1078 & 0.0 & 0.96 \\
I\_SalesDocument & CreationTime & 62862 & 0.0 & 0.97 \\
I\_SalesDocument & CustomerPaymentTerms & 158 & 1.09 & 0.53 \\
I\_SalesDocument & DistributionChannel & 3 & 0.0 & 0.09 \\
I\_SalesDocument & IncotermsClassification & 14 & 0.66 & 0.51 \\
I\_SalesDocument & OrganizationDivision & 1 & 0.0 & 0 \\
I\_SalesDocument & SalesDocument & 500908 & 0.0 & 1.0 \\
I\_SalesDocument & SalesDocumentType & 13 & 0.0 & 0.37 \\
I\_SalesDocument & SalesGroup & 589 & 0.82 & 0.84 \\
I\_SalesDocument & SalesOffice & 45 & 0.01 & 0.18 \\
I\_SalesDocument & SalesOrganization & 34 & 0.0 & 0.58 \\
I\_SalesDocument & ShippingCondition & 56 & 0.01 & 0.55 \\
I\_SalesDocument & TransactionCurrency & 27 & 0.0 & 0.27 \\
I\_SalesDocumentItem & BillToParty & 14789 & 0.0 & 0.78 \\
I\_SalesDocumentItem & IncotermsClassification & 14 & 0.24 & 0.5 \\
I\_SalesDocumentItem & PayerParty & 14755 & 0.0 & 0.78 \\
I\_SalesDocumentItem & Plant & 39 & 0.02 & 0.46 \\
I\_SalesDocumentItem & Product & 187562 & 0.0 & 0.75 \\
I\_SalesDocumentItem & SalesDocument & 501131 & 0.0 & 0.92 \\
I\_SalesDocumentItem & SalesDocumentItem & 720 & 0.0 & 0.59 \\
I\_SalesDocumentItem & SalesDocumentItemCategory & 21 & 0.11 & 0.41 \\
I\_SalesDocumentItem & ShipToParty & 18097 & 0.0 & 0.77 \\
I\_SalesDocumentItem & ShippingPoint & 97 & 0.03 & 0.48 \\
I\_SalesDocumentItem & SoldToParty & 14711 & 0.0 & 0.79 \\
I\_Customer & AddressID & 139611 & 0.0 & 1.0 \\
I\_Customer & Customer & 139609 & 0.0 & 1.0 \\
I\_Address & AddressID & 1788887 & 0.0 & 1.0 \\
I\_Address & AddressRepresentationCode & 1 & 100.0 & 0 \\
I\_Address & Country & 239 & 14.3 & 0.57 \\
I\_Address & Region & 612 & 80.11 & 0.73 \\
\bottomrule
\end{tabular}}
\label{tab:target-field-stats}
\end{table*}

\subsection{Data Extraction and Processing}
\label{sec:data_extraction}
The extracted tables are filtered to contain only the data relevant to this use case.
\begin{itemize}[noitemsep]
\item \texttt{I\_SalesDocument} and \texttt{I\_SalesDocumentItem} are filtered to contain only documents of category sales order
\item Only orders which have been fully processed are included, that is, sales orders which have gone through the entire business process. The goal is to ensure that we are working with the data in its most complete sense and no further changes would be expected.
\item \texttt{I\_Customer} and \texttt{I\_AddrOrgNamePostalAddress} are filtered to contain only the business partners that appear in the aforementioned sales orders and their respective addresses
\item Additionally, the table fields are also filtered to include only those relevant to the use case, which were listed in the previous section
\end{itemize}

After extraction, the tables were joined together to create a flat structure, as described in List. \ref{lst:join-tables}, such that each row in the final table represents one sales order item.
Note that, due to this flat structure, the prediction targets, which are defined on the sales order level, will be repeated across multiple items/rows. This choice was made to allow for the possibility of training a single model to predict all target fields.

\begin{figure}[thp] 
\centering          
\begin{tabular}{c}
\begin{lstlisting}[language=SQL, caption=\textbf{Table join SQL Query:} SQL query used to join the 4 tables together, label={lst:join-tables},basicstyle=\small,frame=single,
  framexleftmargin=10pt,
  framexrightmargin=30pt,
  numbers=left,
  numberstyle=\footnotesize,
  stepnumber=1,
  tabsize=10,
  numbersep=15pt,]
SELECT
        SalesDocumentItem.SalesDocument,
        SalesDocumentItem.SalesDocumentItem,
        SalesDocument.SalesOffice,
        SalesDocument.SalesGroup,
        SalesDocument.CustomerPaymentTerms,
        SalesDocument.ShippingCondition,
        SalesDocumentItem.Plant,
        SalesDocumentItem.ShippingPoint,
        SalesDocument.SalesDocumentType,
        SalesDocument.SalesOrganization,
        SalesDocument.DistributionChannel,
        SalesDocument.OrganizationDivision,
        SalesDocument.BillingCompanyCode,
        SalesDocument.TransactionCurrency,
        SalesDocumentItem.SalesDocumentItemCategory,
        SalesDocumentItem.Product,
        SalesDocumentItem.SoldToParty,
        SoldToPartyAddress.Country as SoldToPartyCountry,
        SoldToPartyAddress.Region as SoldToPartyRegion,
        SalesDocumentItem.ShipToParty,
        ShipToPartyAddress.Country as ShipToCountry,
        ShipToPartyAddress.Region as ShipToPartyRegion,
        SalesDocumentItem.BillToParty,
        BillToPartyAddress.Country as BillToPartyCountry,
        BillToPartyAddress.Region as BillToPartyRegion,
        SalesDocumentItem.PayerParty,
        PayerPartyAddress.Country as PayerCountry,
        PayerPartyAddress.Region as PayerRegion,
        SalesDocument.CreationDate,
        SalesDocument.CreationTime
FROM   I_SalesDocumentItem AS SalesDocumentItem
       INNER JOIN I_SalesDocument AS SalesDocument
              ON SalesDocument.SalesDocument = SalesDocumentItem.SalesDocument
       LEFT JOIN I_Customer AS SoldToPartyTable
              ON SalesDocumentItem.SoldToParty = SoldToPartyTable.Customer
       LEFT JOIN I_Customer AS ShipToPartyTable
              ON SalesDocumentItem.ShipToParty = ShipToPartyTable.Customer
       LEFT JOIN I_Customer AS BillToPartyTable
              ON SalesDocumentItem.BillToParty = BillToPartyTable.Customer
       LEFT JOIN I_Customer AS PayerPartyTable
              ON SalesDocumentItem.PayerParty = PayerPartyTable.Customer
       LEFT JOIN I_AddrOrgNamePostalAddress AS SoldToPartyAddress
              ON SoldToPartyAddress.AddressID = SoldToPartyTable.AddressID
       LEFT JOIN I_AddrOrgNamePostalAddress AS ShipToPartyAddress
              ON ShipToPartyAddress.AddressID = ShipToPartyTable.AddressID
       LEFT JOIN I_AddrOrgNamePostalAddress AS BillToPartyAddress
              ON BillToPartyAddress.AddressID = BillToPartyTable.AddressID
       LEFT JOIN I_AddrOrgNamePostalAddress AS PayerPartyAddress
              ON PayerPartyAddress.AddressID = PayerPartyTable.AddressID
\end{lstlisting}
\end{tabular}
\end{figure}

\begin{figure*}
    \centering
      \begin{subfigure}{0.99\textwidth}
        \includegraphics[width=\textwidth]{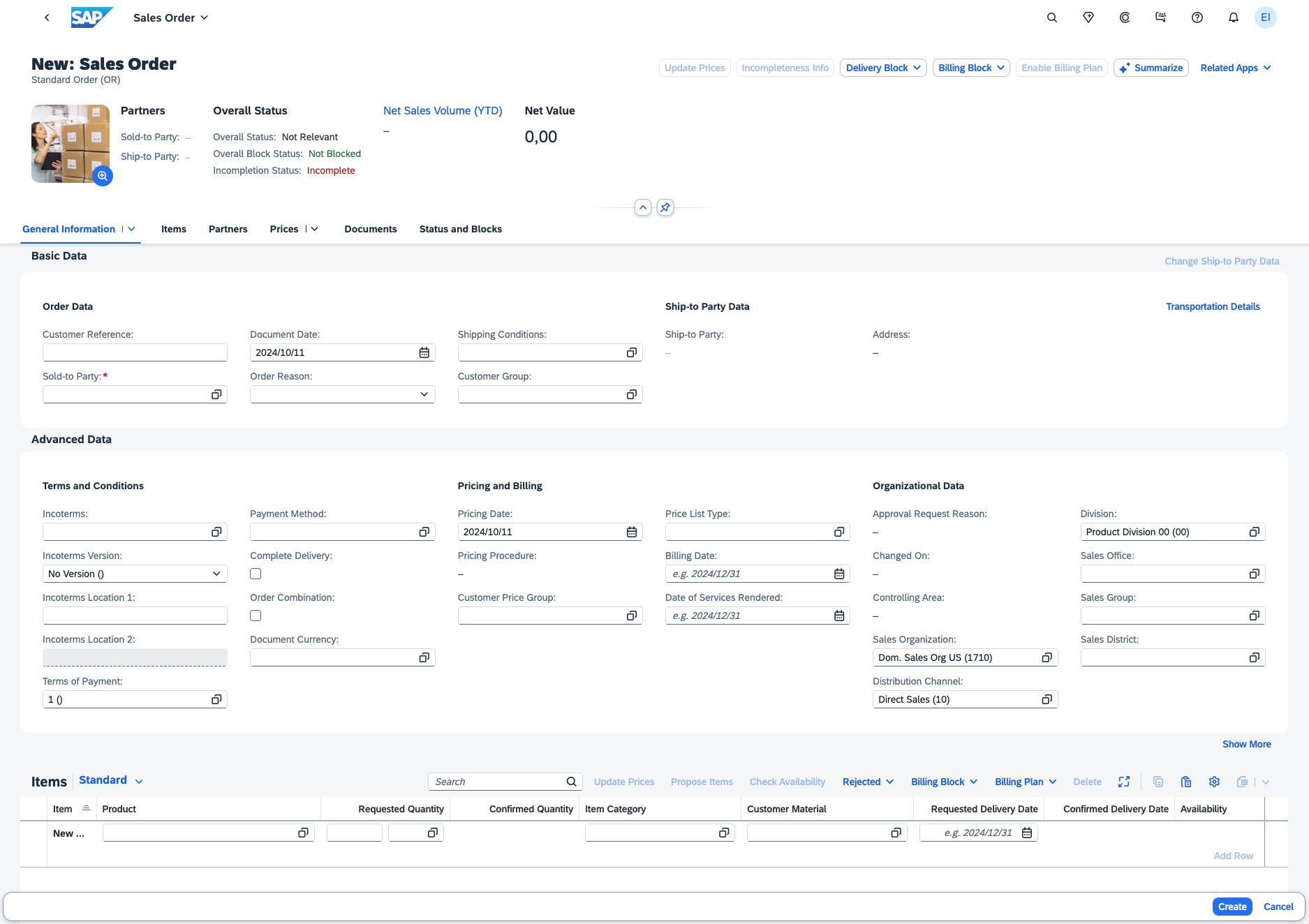}
        \subcaption{Sales Order Creation}
        \label{fig:arm1}
    \end{subfigure}

     \begin{subfigure}{0.99\textwidth}
        \includegraphics[width=\textwidth]{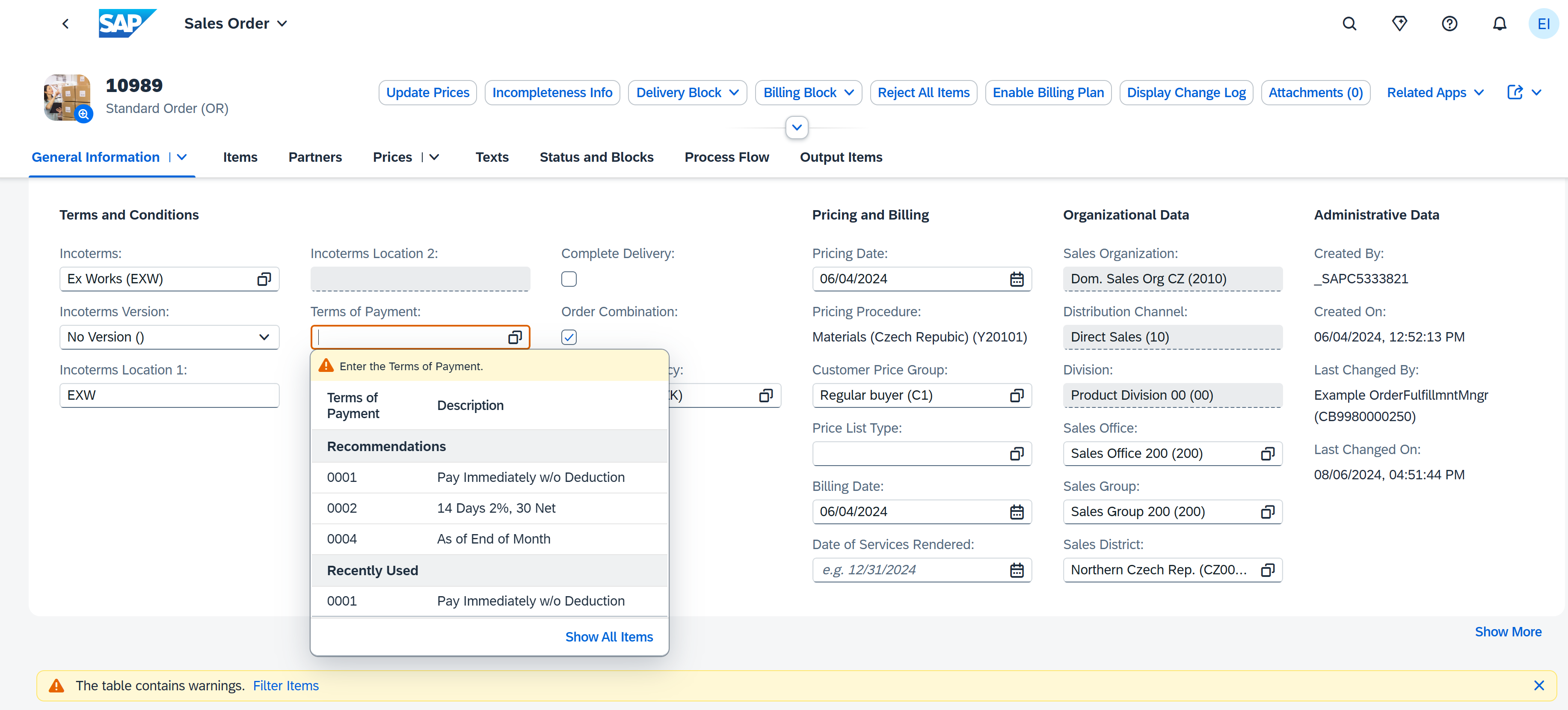}
        \subcaption{Sales Order Editing}
        \label{fig:arm1}
    \end{subfigure}
    \caption{Screenshots of SAP S/4HANA Sales Order User Interface}
    \label{fig:s4hanasalesorderscreenshot}
\end{figure*}

\end{document}